\documentclass[conference]{IEEEtran}
\IEEEoverridecommandlockouts
\usepackage{cite}
\usepackage{amsmath,amssymb,amsfonts}
\usepackage{algorithmic}
\usepackage{graphicx}
\usepackage{textcomp}
\usepackage{xcolor}

\usepackage{amsfonts}
\usepackage{algorithm}
\usepackage{amsmath}
\usepackage{multirow}
\usepackage{array}
\usepackage{amssymb} 

\usepackage{amsthm}

\def\b{\ensuremath\boldsymbol}

\usepackage{url}
\usepackage[algo2e,linesnumbered,lined,boxed,vlined]{algorithm2e}

\usepackage{balance}

\def\BibTeX{{\rm B\kern-.05em{\sc i\kern-.025em b}\kern-.08em
    T\kern-.1667em\lower.7ex\hbox{E}\kern-.125emX}}
\begin{document}
\bstctlcite{IEEEexample:BSTcontrol}


\title{Roweisposes, Including Eigenposes, \\Supervised Eigenposes, and Fisherposes, \\for 3D Action Recognition}



\author{
\IEEEauthorblockN{Benyamin Ghojogh, Fakhri Karray,~\IEEEmembership{Fellow,~IEEE}, Mark Crowley,~\IEEEmembership{Member,~IEEE}
}
\IEEEauthorblockA{Department of Electrical and Computer Engineering, University of Waterloo, Waterloo, ON, Canada \\
Emails: \{bghojogh, karray, mcrowley\}@uwaterloo.ca
}}

\maketitle

\begin{abstract}
Human action recognition is one of the important fields of computer vision and machine learning. Although various methods have been proposed for 3D action recognition, some of which are basic and some use deep learning, the need of basic methods based on generalized eigenvalue problem is sensed for action recognition. This need is especially sensed because of having similar basic methods in the field of face recognition such as eigenfaces and Fisherfaces. In this paper, we propose Roweisposes which uses Roweis discriminant analysis for generalized subspace learning. This method includes Fisherposes, eigenposes, supervised eigenposes, and double supervised eigenposes as its special cases. Roweisposes is a family of infinite number of action recongition methods which learn a discriminative subspace for embedding the body poses. Experiments on the TST, UTKinect, and UCFKinect datasets verify the effectiveness of the proposed method for action recognition. 
\end{abstract}


\section{Introduction}

3D Action recognition uses the information of 3D coordinates of joints \cite{presti20163d}. The methods of action recognition in the literature can be divided into two main categories \cite{khezeli2019time}. The first category deals with the time series of joints and extracts features, such as Dynamic
Time Warping (DTW), from these time series \cite{ghodsi2018simultaneous}.
The second category classifies every frame into some predefined body poses and uses probabilistic approach to recognize the sequence as an action \cite{ghojogh2018fisherposes}.

Some of the action recognition methods are based on convolutional networks \cite{ji20123d,wang2018action}. 
Some other methods, however, have a manifold-based or subspace-based approaches. In these approaches, the actions lie on subspaces of data \cite{hong2019variant}. Riemannian manifolds \cite{devanne20143} and Grassmann manifolds \cite{slama2015accurate} have been extensively used for action recognition. 
Some of the methods embed actions in sparse manifolds \cite{mohammadzade2020sparsness} to make use of betting on sparsity principle \cite{hastie2009elements}. 

Some of the action recognition methods use basic subspace learning methods for recognizing poses or actions. For example, histograms of 3D joint locations (HOJ3D) utilizes Linear Discriminant Analysis (LDA) which is equivalent to Fisher Discriminant Analysis (FDA) \cite{ghojogh2019linear}. LDA is also used in \cite{mokari2020recognizing} for recognizing involuntary actions. 
Fisherposes is a method which uses FDA \cite{hastie2009elements,fisher1936use} for embedding the body poses and their recognition \cite{ghojogh2018fisherposes}.
The reader can refer to \cite{ghojogh2018fisherposes} and the references therein to see papers in action recognition which use LDA or FDA. FDA is very useful for pose embedding and action recognition because it maximizes and minimizes the inter-class and intra-class scatters of body poses, respectively \cite{ghojogh2019fisher}. 

There exist many subspace learning methods which are based on generalized eigenvalue problem \cite{ghojogh2019eigenvalue}.
Some examples of these methods are Principal Component Analysis (PCA) \cite{jolliffe2011principal,ghojogh2019unsupervised}, Supervised PCA (SPCA) \cite{barshan2011supervised}, and FDA \cite{ghojogh2019fisher}. The subspace of PCA maximizes the scatter of projected data while SPCA has a subspace which maximizes the statistical dependence of projected data and the class labels. 
Roweis Discriminant Analysis (RDA) \cite{ghojogh2019roweis,ghojogh2020roweis} is a family of infinite number of subspace learning methods based on generalized eigenvalue problem. It generalizes PCA, SPCA, and FDA and also includes Double Supervised Discriminant Analysis (DSDA) \cite{ghojogh2020roweis}.  

Basic subspace learning methods, such as PCA, SPCA, and FDA, can be used for 3D action recognition for embedding poses and classifying actions. There have appeared different face recognition methods using basic subspace learning approaches \cite{abate20072d}. Some examples are eigenfaces \cite{turk1991face}, Fisherfaces \cite{belhumeur1997eigenfaces}, supervised eigenfaces \cite{barshan2011supervised,ghojogh2019unsupervised}, and double supervised eigenfaces \cite{ghojogh2020roweis}. 
Similar need to methods based on basic subspace learning is sensed for action recognition literature. 

In this paper, we propose \textit{Roweisposes} for action recognition. This method is based on basic subspace learning approaches which make use of generalized eigenvalue problem \cite{ghojogh2019eigenvalue}. This method, which uses RDA \cite{ghojogh2020roweis}, generalizes the Fisherposes method \cite{ghojogh2018fisherposes}. Some of the special cases of Roweisposes are based on PCA, SPCA, and DSDA and we name them \textit{eigenposes}, \textit{supervised eigenposes}, and \textit{double supervised eigenposes}, respectively. 
The Roweisposes method includes infinite number of subspace learning methods for embedding the body poses useful for action recognition. 

The remainder of this paper is organized as follows. 
Section \ref{section_Fisherposes} reviews FDA and how it can be restated using the total scatter. Moreover, the Fisherposes method is reviewed in this section. Section \ref{section_RDA} reviews the theory of PCA and SPCA as well as RDA and the Roweis map. The Roweisposes method is proposed in Section \ref{section_Roweisposes} where pre-processing, training the Roweisposes subspace, pose recognition, windowing, training hidden Markov model, and action recognition are detailed. The experiments are reported in Section \ref{section_experiments} to show the effectiveness of the proposed method. Finally, Section \ref{section_conclusion} concludes the paper and enumerates the future work. 

\section{Review of Fisherposes}\label{section_Fisherposes}

\subsection{Fisher Discriminant Analysis}

Let $\{\b{x}_i\}_{i=1}^n$ be a dataset of sample size $n$ and dimensionality $d$. We denote the number of classes by $c$, the sample size of $j$-th class by $n_j$, and the $i$-th instance of the $j$-th class by $\b{x}_i^{(j)}$. 
FDA \cite{hastie2009elements,ghojogh2019fisher}, first proposed in \cite{fisher1936use}, maximizes the inter-class (between) scatter and minimizes the intra-class (within) scatter of the projected data:
\begin{equation}\label{equation_optimization_FDA}
\begin{aligned}
& \underset{\b{U}}{\text{maximize}}
& & \textbf{tr}(\b{U}^\top \b{S}_B\, \b{U}), \\
& \text{subject to}
& & \b{U}^\top \b{S}_W\, \b{U} = \b{I},
\end{aligned}
\end{equation}
where $\b{S}_B$ and $\b{S}_W$ are the between and within scatters, respectively, defined as:
\begin{align}
&\mathbb{R}^{d \times d} \ni \b{S}_B := \sum_{j=1}^c n_j (\b{\mu}_j - \b{\mu}) (\b{\mu}_j - \b{\mu})^\top, \label{equation_between_scatter} \\
&\mathbb{R}^{d \times d} \ni \b{S}_W := \sum_{j=1}^c \sum_{i=1}^{n_j} (\b{x}_i^{(j)} - \b{\mu}_j) (\b{x}_i^{(j)} - \b{\mu}_j)^\top, \label{equation_within_scatter}
\end{align}
and the total mean and the mean of the $j$-th class are $\mathbb{R}^{d} \ni \b{\mu} := (1/n) \sum_{i=1}^{n} \b{x}_i$ and $\mathbb{R}^{d} \ni \b{\mu}_j := (1/n_j) \sum_{i=1}^{n_j} \b{x}_i^{(j)}$. 
The solution to Eq. (\ref{equation_optimization_FDA}) is the generalized eigenvalue problem $(\b{S}_B, \b{S}_W)$ \cite{ghojogh2019eigenvalue}. 

The total scatter can be considered as the summation of the between and within scatters \cite{ye2007least}:
\begin{align}\label{equation_S_T_as_sum_of_scatters}
\b{S}_T = \b{S}_B + \b{S}_W \implies \b{S}_B = \b{S}_T - \b{S}_W,
\end{align}
where the covariance matrix, or the total scatter, is defined as:
\begin{equation}\label{equation_total_scatter}
\begin{aligned}
\mathbb{R}^{n \times n} \ni \b{S}_T &:= \sum_{i=1}^n (\b{x}_i - \b{\mu})(\b{x}_i - \b{\mu})^\top \\
&= \b{X} \b{H} \b{H} \b{X}^\top \overset{(a)}{=} \b{X} \b{H} \b{X}^\top, 
\end{aligned}
\end{equation}
where $\b{H}$ is the centering matrix:
\begin{align}
\mathbb{R}^{n \times n} \ni \b{H} := \b{I} - (1/n) \b{1}\b{1}^\top,
\end{align}
with $\b{I}$ as identity matrix and $\b{1}$ as the vectors of ones. The $(a)$ is because the centering matrix is symmetric and idempotent. 

Hence, the Fisher criterion can be written as:
\begin{align}\label{equation_optimization_FDA_criterion_with_S_T}
&\frac{\textbf{tr}(\b{U}^\top \b{S}_B\, \b{U})}{\textbf{tr}(\b{U}^\top \b{S}_W\, \b{U})} = \frac{\textbf{tr}(\b{U}^\top \b{S}_T\, \b{U})}{\textbf{tr}(\b{U}^\top \b{S}_W\, \b{U})} - 1. 
\end{align}
The $-1$ is a constant and can be dropped in the optimization problem because the variable $\b{U}$ and not the objective is the goal; therefore, the optimization in FDA can be expressed as:
\begin{equation}\label{equation_optimization_FDA_with_S_T}
\begin{aligned}
& \underset{\b{U}}{\text{maximize}}
& & \textbf{tr}(\b{U}^\top \b{S}_T\, \b{U}), \\
& \text{subject to}
& & \b{U}^\top \b{S}_W\, \b{U} = \b{I},
\end{aligned}
\end{equation}
whose solution is the generalized eigenvalue problem $(\b{S}_T, \b{S}_W)$ \cite{ghojogh2019eigenvalue}. 
Hence, the FDA subspace is spanned by the eigenvectors of this generalized eigenvalue problem. 

\subsection{Fisherposes}

The Fisherposes method  \cite{ghojogh2018fisherposes} is an action recognition approach which uses 3D skeletal data as input and constructs a Fisher subspace, for discrimination of body poses, using FDA. 
Instead of using the raw data, it applied some pre-processing on the 3D data. These pre-processing steps include skeleton alignment by translating the hip joint to the origin and aligning the shoulders to cancel the orientation of body. Moreover, the scales of skeletons are removed and some informative joints are selected amongst all the available joints. 

After the pre-processing step, different body poses are selected out of the dataset where every action can be decomposed into a sequence of some of these poses.
The body poses are considered as classes and the instances of a body pose are used as the data of that class. The information of joints in a body pose are concatenated to form a vector. Using these data vectors, the FDA subspace is trained to discriminate the body poses of an action recognition dataset. 

Using Euclidean distance of the projected frame onto the FDA subspace from the projection of training data, the pose of a frame is recognized. 
Windowing is also applied to eliminate the frames which do not belong to any of the poses well enough. The distance of projection onto the subspace is used as a criterion for windowing. 
Finally a Hidden Markov Model (HMM) \cite{ghojogh2019hidden} is used to learn the sequences of  recognized poses as different actions. For some datasets in which some actions contain similar poses without consideration of movement of body, histogram of trajectories is also used to discriminate those actions. 

\section{Roweis Discriminant Analysis}\label{section_RDA}

\subsection{PCA and SPCA}

PCA \cite{jolliffe2011principal,ghojogh2019unsupervised} finds a subspace which maximizes the scatter of projected data. Its optimization is:
\begin{equation}\label{equation_optimization_PCA}
\begin{aligned}
& \underset{\b{U}}{\text{maximize}}
& & \textbf{tr}(\b{U}^\top \b{S}_T\, \b{U}), \\
& \text{subject to}
& & \b{U}^\top \b{U} = \b{I},
\end{aligned}
\end{equation}
where $\b{S}_T \in \mathbb{R}^{n \times n}$ is the total scatter defined in Eq. (\ref{equation_total_scatter}).
The solution to Eq. (\ref{equation_optimization_PCA}) is the eigenvalue problem for $\b{S}_T$ \cite{ghojogh2019eigenvalue}.
Hence, PCA subspace is spanned by the eigenvectors of total scatter. 

SPCA \cite{barshan2011supervised} uses the empirical estimation of the Hilbert-Schmidt Independence Criterion (HSIC) \cite{gretton2005measuring}:
\begin{align}\label{equation_HSIC}
\text{HSIC} := \frac{1}{(n-1)^2}\, \textbf{tr}(\b{K}_1 \b{H}\b{K}_2 \b{H}),
\end{align}
where $\b{K}_1$ and $\b{K}_2$ are the kernels over the first and second random variables, respectively. The idea of HSIC is to measure the dependence of two random variables by calculating the correlation of their pulled values to the Hilbert space. 
SPCA uses HSIC and to maximize the dependence of the projected data $\b{U}^\top \b{X}$ and the labels $\b{Y}$. It uses the linear kernel for the projected data $\b{K}_1 = (\b{U}^\top \b{X})^\top (\b{U}^\top \b{X}) = \b{X}^\top \b{U} \b{U}^\top \b{X}$ and an arbitrary valid kernel for the labels $\b{K}_2=\b{K}_y$. Therefore, the scaled Eq. (\ref{equation_HSIC}) in SPCA is $\textbf{tr}(\b{X}^\top \b{U} \b{U}^\top \b{X} \b{H}\b{K}_y \b{H}) \overset{(a)}{=} \textbf{tr}(\b{U}^\top \b{X} \b{H} \b{K}_y \b{H} \b{X}^\top \b{U})$ where $(a)$ is because of the cyclic property of the trace. 
The optimization problem in SPCA is:
\begin{equation}\label{equation_optimization_SPCA}
\begin{aligned}
& \underset{\b{U}}{\text{maximize}}
& & \textbf{tr}(\b{U}^\top \b{X} \b{H} \b{K}_y \b{H} \b{X}^\top \b{U}), \\
& \text{subject to}
& & \b{U}^\top \b{U} = \b{I},
\end{aligned}
\end{equation}
where $\b{K}_y$ is the kernel matrix over the labels of data, either for classification or regression. 
The solution to Eq. (\ref{equation_optimization_SPCA}) is the eigenvalue problem for $\b{X}\b{H}\b{K}_y\b{H}\b{X}^\top$ \cite{ghojogh2019eigenvalue}.
Hence, the SPCA directions are the eigenvectors of $\b{X}\b{H}\b{K}_y\b{H}\b{X}^\top$. 

\subsection{RDA and Roweis Map}

Comparing Eqs. (\ref{equation_optimization_FDA_with_S_T}), (\ref{equation_optimization_PCA}), and (\ref{equation_optimization_SPCA}) shows that they follow a general form of optimization:
\begin{equation}\label{equation_optimization_generalForm_generalized_eigenvalue}
\begin{aligned}
& \underset{\b{U}}{\text{maximize}}
& & \textbf{tr}(\b{U}^\top \b{S}_1\, \b{U}), \\
& \text{subject to}
& & \b{U}^\top \b{S}_2\, \b{U} = \b{I},
\end{aligned}
\end{equation}
whose solution is the generalized eigenvalue problem $(\b{S}_1, \b{S}_2)$ \cite{ghojogh2019eigenvalue}. 

\begin{figure}[!t]
\centering
\includegraphics[width=2.5in]{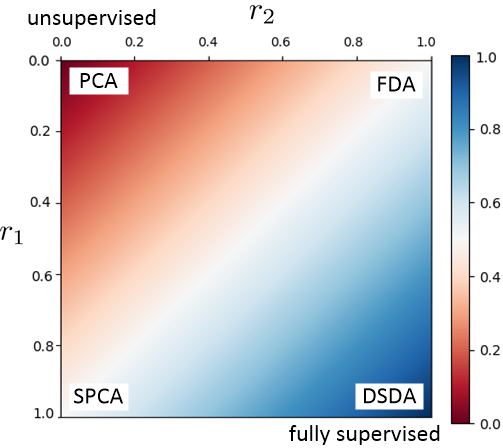}
\caption{The Roweis map where the color indicates the supervision level. Credit of image is for \cite{ghojogh2019roweis,ghojogh2020roweis}.} 
\label{figure_Roweis_map}
\end{figure}

Following this general form, RDA \cite{ghojogh2019roweis,ghojogh2020roweis} solves this optimization problem: 
\begin{equation}\label{equation_optimization_RDA}
\begin{aligned}
& \underset{\b{U}}{\text{maximize}}
& & \textbf{tr}(\b{U}^\top \b{R}_1\, \b{U}), \\
& \text{subject to}
& & \b{U}^\top \b{R}_2\, \b{U} = \b{I},
\end{aligned}
\end{equation}
where $\b{R}_1$ and $\b{R}_2$ are the first and second Roweis matrices which are:
\begin{align}
&\mathbb{R}^{d \times d} \ni \b{R}_1 := \b{X} \b{H} \b{P} \b{H} \b{X}^\top, \label{equation_R1} \\
&\mathbb{R}^{d \times d} \ni \b{R}_2 := r_2\, \b{S}_W + (1 - r_2)\, \b{I}, \label{equation_R2}
\end{align}
respectively, where:
\begin{align}\label{equation_P}
\mathbb{R}^{n \times n} \ni \b{P} := r_1\, \b{K}_y + (1 - r_1)\, \b{I}.
\end{align}
The variables $r_1 \in [0,1]$ and $r_2 \in [0,1]$ are the first and second Roweis factors. 
Changing the Roweis factors gives a Roweis map including infinite number of subspace learning methods. This map, as well as the supervision level $s = (r_1 + r_2) / 2$, is illustrated in Fig. \ref{figure_Roweis_map}. As this figure shows, four extreme special cases of RDA are PCA (with $r_1=0,r_2=0$) \cite{jolliffe2011principal,ghojogh2019unsupervised}, SPCA (with $r_1=1,r_2=0$) \cite{barshan2011supervised}, FDA (with $r_1=0,r_2=1$) \cite{fisher1936use,ghojogh2019fisher}, and Double Supervised Discriminant Analysis (DSDA) (with $r_1=1,r_2=1$) \cite{ghojogh2019roweis,ghojogh2020roweis}. 
The RDA subspace is spanned by the eigenvectors of the generalized eigenvalue problem $(\b{R}_1, \b{R}_2)$ \cite{ghojogh2019eigenvalue}.

\section{Roweisposes for 3D Action Recognition}\label{section_Roweisposes}

We propose Roweisposes, for 3D action recognition, which generalizes the Fisherposes method (with $r_1=0,r_2=1$). In addition to this generalization, it also includes new methods for action recognition, namely eigenposes (with $r_1=0,r_2=0$), supervised eigenposes (with $r_1=1,r_2=0$), and double supervised eigenposes (with $r_1=1,r_2=1$). 
The Roweisposes method uses 3D skeletal information of frames and its subspace discriminates the body poses for an action dataset. 

\subsection{Pre-processing and Data Preparation}

Inspired by Fisherposes \cite{ghojogh2018fisherposes}, some pre-processing steps are applied to the joints. These steps are translating hip to the origin, shoulder alignment, removing scale of joints, and selecting the most informative joints. The details of these steps can be found in \cite{ghojogh2018fisherposes} and we do not repeat for the sake of brevity.
Different body poses are selected from the actions, either manually \cite{ghojogh2018fisherposes} or automatically \cite{ghojogh2017automatic}. 
We take the body poses as the classes to be discriminated by the RDA subspace. The data for training this subspace are created by vectorizing the information of joints in the frame indicating the body pose. In other words, if the number of selected joints is denoted by $J$ and the 3D coordinate of the $l$-th joint is denoted by $[x_l, y_l, z_l]^\top$, the vector of body pose in a frame is vectorized as:
\begin{align}
\mathbb{R}^{3J} \ni \b{x} = [x_1, \dots, x_J, y_1, \dots, y_J, z_1, \dots, z_J]^\top,
\end{align}
hence, $d = 3J$.

\subsection{Training the Roweisposes Subspace}

Using the vectorized frames of body poses and their pose labels, we train an RDA subspace for discriminating the poses. This subspace is named Roweisposes. 
The Roweisposes subspace is spanned by the eigenvectors of the generalized eigenvalue problem $(\b{R}_1, \b{R}_2)$ according to Eq. (\ref{equation_optimization_RDA}). 

After training this subspace, every training body frame, indexed by $i$, is projected onto this subspace:
\begin{align}
\mathbb{R}^p \ni \widehat{\b{x}}_i = \b{U}^\top \b{x}_i,
\end{align}
where $\widehat{\b{x}}_i$ denotes the projected data, $p$ is the dimensionality of subspace, and $\b{U} \in \mathbb{R}^{d \times p}$ is the projection matrix whose columns are the $p$ leading eigenvectors of the generalized eigenvalue problem $(\b{R}_1, \b{R}_2)$. 

\subsection{Pose Recognition}

The body pose of an unknown train/test frame, $\b{x}_t$, is then recognized using Euclidean distances from the means of projected classes onto the Roweisposes subspace. The minimum distance determines the body pose:
\begin{align}\label{equation_pose_recognition}
\text{pose}_t := \arg \min_{j=1}^c \|\b{x}_t - \widehat{\b{\mu}}_j \|_2,
\end{align}
where $\|.\|_2$ denotes the $\ell_2$ norm, $c$ is the number of classes (poses), and $\widehat{\b{\mu}}_j$ is the mean of the projection of the $j$-th class onto this subspace:
\begin{align}\label{equation_mean_projection}
\mathbb{R}^p \ni \widehat{\b{\mu}}_j := \frac{1}{n_j} \sum_{i=1}^{n_j} \widehat{\b{x}}_i^{(j)}.
\end{align}

\subsection{Windowing}

Some of the frames are the middle frames as transition between two poses. These frames are better to be removed from the sequence which is going to be fed to HMM. This helps HMM learn the sequences of frames more accurately. However, this elimination of frames may make some sequences very short. In order to have a trade-off between appropriate sequence length and the not-accurate frame removal, we use windowing. In windowing, we do not let all the frames within the window be removed. Similar approach of windowing in Fisherposes is used here (see {\cite[Algorithm 1]{ghojogh2018fisherposes}}). 

\subsection{Training HMM model}

The next step is to train an HMM model \cite{ghojogh2019hidden} using the recognized poses of the training body frames using the Roweisposes subspace. 
The training frames of every sequence are projected onto the subspace and their poses are recognized using Eq. (\ref{equation_pose_recognition}). Sequences of the recognized body frames are fed to HMM and expectation maximization and the Baum-Welch algorithm \cite{baum1970maximization} are employed to train the model. 

\subsection{Action Recognition}

In the test phase, the frames of the test sequence are projected onto the Roweisposes subspace and their poses are recognized using Eq. (\ref{equation_pose_recognition}). The sequence of recognized frames are then fed to the HMM. 
For the test phase in HMM, we use the Viterbi algorithm to determine the action of the test sequence \cite{jurafsky2019speech}. The action with the highest likelihood is determined as the action of that test sequence \cite{ghojogh2019hidden}. 

\section{Experiments}\label{section_experiments}

\subsection{Datasets}

For validating the effectiveness of the proposed Roweisposes, we used three publicly available datasets following the paper \cite{ghojogh2018fisherposes}. 
In the following, we introduce the characeristics of these datasets.

\subsubsection{TST Dataset}

The first dataset is the TST fall detection \cite{gasparrini2015proposal} which includes two categories of normal and fall actions. The normal actions are sitting, grasping, walking, and lying down, and the fall actions are falling front, back, side, and falling backward which ends up sitting. The number of subjects performing the actions are 11.  

\subsubsection{UTKinect Dataset}

The UTKinect dataset \cite{xia2012view} contains 10 actions which are walking, sitting down, standing up, picking up, carrying, throwing, pushing, pulling, waving, and clapping hands. This dataset has 10 subjects performing these actions.

\subsubsection{UCFKinect Dataset}

The UCFKinect dataset \cite{ellis2013exploring} includes 16 actions which are balancing, climbing ladder, ducking, hopping, kicking, leaping, punching, running, stepping back, stepping front, stepping left, stepping right, twisting left, twisting right, and vaulting. The number of subjects in this dataset is 16. 

\subsection{Experimental Setup}

The selected joints in the skeletal data of the three datasets were followed by the selections in the Fisherposes method. The reader can refer to {\cite[Fig. 2]{ghojogh2018fisherposes}} for viewing the selected joints in these datasets. 
The selected poses for the three datasets follow {\cite[Fig. 5]{ghojogh2018fisherposes}}. 
For the experiments, we used leave-one-person-out cross validation. The hyperparameters for windowing and histogram of trajectories were determined according the the paper \cite{ghojogh2018fisherposes}. 
Euclidean distance was used for recognizing the body poses after projection onto the Roweisposes subspace. Employing the regularized Mahalanobis distance is deferred to future work (see Section \ref{section_conclusion}). 

\begin{table}[!t]
\caption{Comparison of the special cases of Rowiesposes with other action recognition methods. The rates are average accuracy.}
\label{table_accuracy}
\centering
\scalebox{0.95}{    
\begin{tabular}{l || c | c | c }
& TST & UTKinect & UCFKinect   \\
\hline
\hline
Grassmann manifold \cite{slama2015accurate} & -- & 88.50\% & 97.91\% \\
HOJ3D \cite{xia2012view} & 70.83\% & 90.92\% & -- \\
kinetic energy \cite{shan20143d} & 84.09\% & -- & -- \\
CRR \cite{tabejamaat2020contributive} & 89.39\% & 95.98\% & -- \\
Riemannian manifold \cite{devanne20143} & -- & 91.50\% & -- \\
MDTW \cite{ghodsi2018simultaneous} & 92.30\% & 96.80\% & 97.90\% \\
Sparseness Embedding \cite{mohammadzade2020sparsness} & 94.27\% & 92.00\% & -- \\
\hline
Roweisposes ($r_1=0, r_2=0$) & 81.44\% & 38.50\% & 87.19\% \\
Roweisposes ($r_1=0, r_2=1$) \cite{ghojogh2018fisherposes} & 76.14\% & 82.50\% & 79.22\% \\
Roweisposes ($r_1=1, r_2=0$) & 82.20\% & 70.50\% & 86.80\% \\
Roweisposes ($r_1=1, r_2=1$) & 76.52\% & 79.00\% & 71.72\% \\
\hline
Roweisposes ($r_1=0, r_2=0.5$) & 79.17\% & 83.50\% & 80.02\% \\
Roweisposes ($r_1=1, r_2=0.5$) & 80.68\% & 82.50\% & 86.25\% \\
Roweisposes ($r_1=0.5, r_2=0$) & 79.92\% & 41.00\% & 88.36\% \\
Roweisposes ($r_1=0.5, r_2=1$) & 80.30\% & 82.50\% & 69.45\% \\
Roweisposes ($r_1=0.5, r_2=0.5$) & 81.82\% & 80.50\% & 86.25\% \\
\hline
\hline
\end{tabular}%
}
\end{table}

\subsection{Performance of Roweisposes}

\begin{figure*}[!t]
\centering
\includegraphics[width=6.5in]{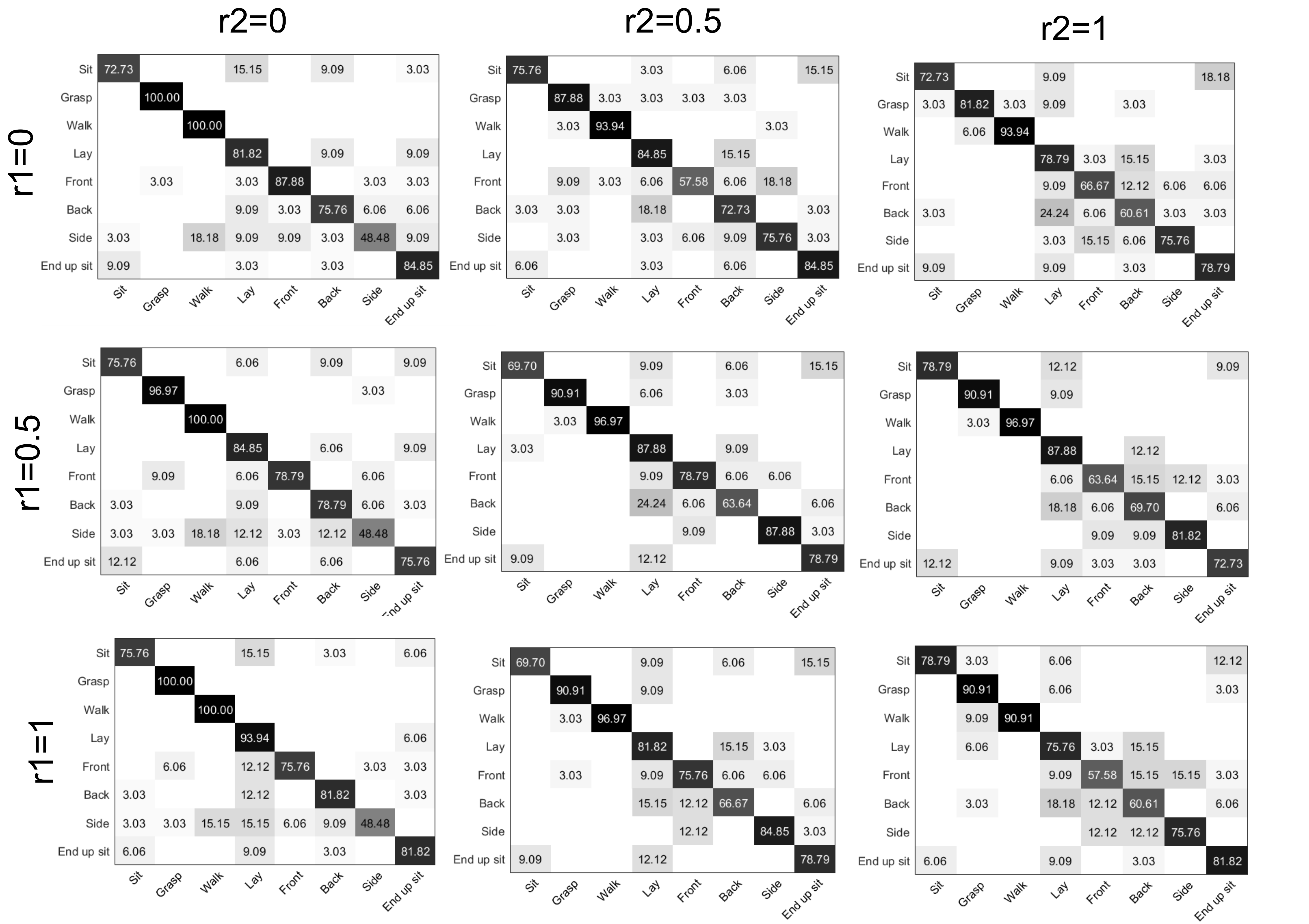}
\caption{Confusion matrices of the Roweisposes performance on the TST dataset.} 
\label{figure_confusion_matrices}
\end{figure*}

The average of accuracies over the cross validation folds are reported in Table \ref{table_accuracy}. The first part of table reports the related work and the state-of-the-art performances on the TST, UTKinect, and UCFKinect datasets. 
The second part of table contains the performances of the four extreme cases of Roweisposes which are eigenposes (with $r_1=0,r_2=0$), Fisherposes (with $r_1=0,r_2=1$), supervised eigenposes (with $r_1=1,r_2=0$), and double supervised eigenposes (with $r_1=1,r_2=1$). 
the thirs part of table reports the performance of some of the middle special cases of Roweisposes in the Roweis map. These cases are ($r_1=0,r_2=0.5$), ($r_1=1,r_2=0.5$), ($r_1=0.5,r_2=0$), ($r_1=0.5,r_2=1$), and ($r_1=0.5,r_2=0.5$).

Regarding the comparison of the special cases of Roweisposes against each other, we see that, except some exceptions, higher supervision level mostly results in better performance. In other words, usually, Fisherposes and supervised eigenposes have superior or comparable performance than eigenposes. This is especially true for the UTKinect dataset while in the other datasets, the performances are comparable. This is expected because more level of supervision make more use of the labels and thus improves the recognition by learning a more discriminative subspace for poses. The performance of double supervised eigenposes is not necessarily much better than the other cases. This fact has been shown to be also true for facial image and the MNIST dataset in the paper \cite{ghojogh2020roweis}. That paper shows that only in some regression problems, which is not the case study of this paper, DSDA outperforms other cases. 

The performance of the cases $r_1=0,r_2=1$ is slightly different that reported for Fisherposes in paper \cite{ghojogh2018fisherposes}. This has two small reasons. The first reason is that FDA as a special case of RDA uses the optimization problem (\ref{equation_optimization_FDA_with_S_T}) but the Fisherposes method in paper \cite{ghojogh2018fisherposes} uses problem (\ref{equation_optimization_FDA}). 
The second reason is that this paper uses Euclidean distance in Eq. (\ref{equation_pose_recognition}) for simplicity of the method and elimination of a hyperparameter. The paper \cite{ghojogh2018fisherposes} uses a regularized Mahalanobis distance instead. We have deferred trying the Roweisposes method with Mahalanobis distance to the future work (see Section \ref{section_conclusion}). 
The middle special cases of Roweisposes also show that the performance of this method is almost stable in most of the cases of Roweisposes in the Roweis map. 

\begin{figure*}[!t]
\centering
\includegraphics[width=5in]{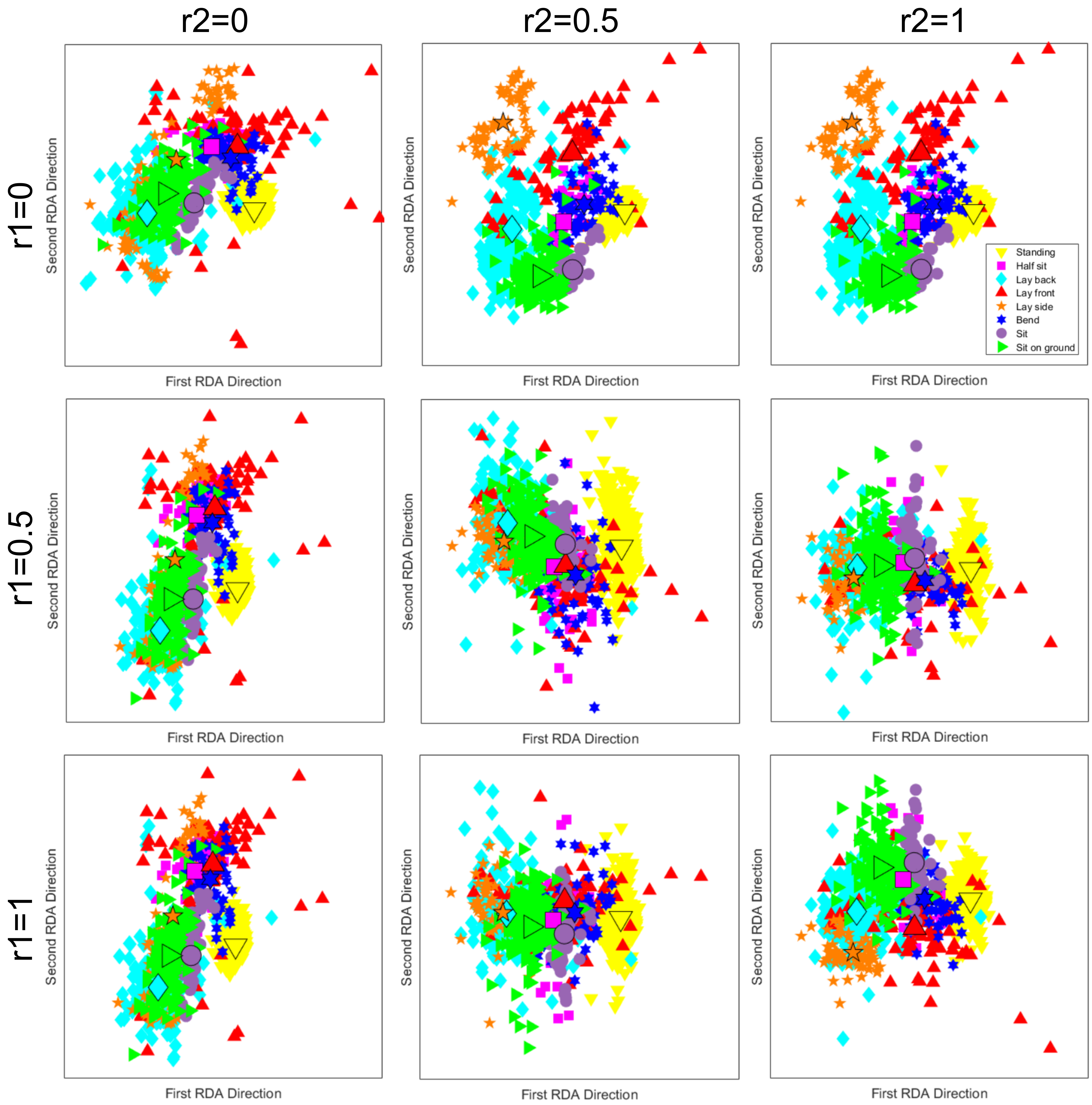}
\caption{The embedded body poses of the TST dataset in the two leading Roweisposes directions. The larger shapes indicate the means of projections of poses onto the subspace (see Eq. (\ref{equation_mean_projection})).} 
\label{figure_subspaces_TST}
\end{figure*}

The related work and state-of-the-art methods reported in Table \ref{table_accuracy} are the Grassmann manifold \cite{slama2015accurate}, HOJ3D \cite{xia2012view}, kinetic energy \cite{shan20143d},  Contributive Representation based Reconstruction (CRR) \cite{tabejamaat2020contributive}, Riemannian manifold \cite{devanne20143}, MDTW \cite{ghodsi2018simultaneous}, Multidimensional DTW (MDTW) \cite{ghodsi2018simultaneous}, and Sparseness embedding \cite{mohammadzade2020sparsness}. 
On TST data, Roweisposes outperforms HOJ3D and has comparable performance with the kinetic energy method. Although the Rowesiposes method does not outperform the state of the art in other cases, this does not question the value of the proposed method because it proposes infinite number of action recognition methods using basic subspace learning methods based on generalized eigenvalue problem. The rich literature of action recognition has a lack for such fundamental methods; although these basic methods have been vastly used in other fields such as face recognition \cite{abate20072d}. 

\subsection{Confusion Matrices}

Figure \ref{figure_confusion_matrices} depicts the confusion matrices of the performance of Roweisposes in nine special cases of the Roweis map. For the sake of brevity, we only report the matrices for TST dataset. As this figure shows, the performance of Roweisposes is stable and acceptable along different parts of the map. However, some actions are recognized more differently in different cases. 
For example, falling side is recognized better by increasing $r_2$. This makes sense because of using more information of labels in the kernel over labels (see Eq. (\ref{equation_P})). In larger values of $r_2$, this action is confused with similar actions such as falling front, falling back, and falling ending up sitting. An example for the impact of increasing $r_1$ is the grasping action. In most of the cases, increasing $r_1$ improves the the performance of Roweisposes for recognizing this action which is expected because of more use of labels in the within scatter (see Eq. (\ref{equation_R2})). 

\subsection{Comparison of Subspaces}

The embedding of body poses in the two leading Roweisposes directions are depicted in Fig. \ref{figure_subspaces_TST}. For brevity, we only show the embedding of TST dataset. This figure shows the embeddings along nine different cases in the Roweis map. 
As this figure shows, poses are more more confused in the smaller supervision level, i.e., smaller values of $r_1$ and $r_2$. This is expected because more supervision level uses the information of pose labels more. For instance, the pose laying side is confused when $r_2=0$. Increasing $r_2$ discriminates this pose much better. The poses laying back and sitting on ground are also better separated when $r_2$ is increased. As an example for effect of $r_1$, in $r_2=1$, increasing $r_1$ from $0.5$ to $1$ has separated the poses laying side and laying back more. 

\section{Conclusion and Future Work}\label{section_conclusion}

In this paper, we proposed the Roweisposes method which generalizes Fisherposes. This method includes infinite number of action methods which learn a discriminative subspace for embedding body poses using the generalized eigenvalue problem. Four special cases of Roweisposes are Fisherposes, eigenposes, supervised eigenposes, and double supervised eigenposes. Compared to the complicated new action recognition methods, the need to this basic action recognition method is especially sensed when similar basic approaches have been proposed for face recognition such as Fisherfaces, eigenfaces, supervised eigenfaces, and double supervised eigenfaces. 

At least two possibilities for future direction exist to improve the Roweisposes method. The first improvement is to use Mahalanobis distance instead of Euclidean distance for recognizing the body pose after projection onto the Roweisposes subspace. Tuning the parameters of regularized Mahalanobis distance \cite{ghojogh2018fisherposes} may improve the accuracy of action recognition. The other future direction is to sampling the training frames for learning the body poses automatic rather than manual selection \cite{ghojogh2017automatic}.

\bibliographystyle{IEEEtran}
\bibliography{references}

\end{document}